\documentclass[submission,copyright,creativecommons]{eptcs}

\vspace{-1mm}
\usepackage{titlesec}
\titlespacing*{\section}
{0pt}{2ex}{2ex}
\titlespacing*{\subsection}
{0pt}{1.5ex}{1.5ex}
\titlespacing*{\subsubsection}
{0pt}{1.2ex}{1.2ex}

\newcommand{\sumijn}{\sum_{\substack{i \in 1...n\\ j \in 1...n}}}
\newcommand{\sumijx}{\sum_{\substack{i,j \in 1...|X|}}}

\newcommand{\adistij}{\|A_i - A_j\|}

\newcommand{\db}{\mathbf{D}}
\newcommand{\cb}{\mathbf{C}}

\newcommand{\rl}{\mathbb{R}}
\newcommand{\rlp}{\mathbb{R}_{\geq 0}}

\newcommand{\cl}{\mathcal{C}}

\newcommand{\bx}{\mathbf{X}}

\newcommand{\fuzz}{\mathbf{F}}

\newcommand{\lij}[1]{l_{{#1}_{ij}}}
\newcommand{\cij}[1]{c_{{#1}_{ij}}}
\newcommand{\eij}[1]{e_{{#1}_{ij}}}

\newcommand{\cfaij}{\cij{F(a)}}
\newcommand{\efaij}{\eij{F(a)}}

\newcommand{\pmet}{\mathcal{\mathbf{PMet}}}
\newcommand{\pmetinj}{\pmet_{\mathit{inj}}}
\newcommand{\pmetbij}{\pmet_{\mathit{bij}}}
\newcommand{\pmetsur}{\pmet_{\mathit{sur}}}
\newcommand{\pmetisom}{\pmet_{\mathit{isom}}}

\newcommand{\cv}{\mathbf{Cov}}
\newcommand{\cvs}{\cv}
\newcommand{\cvsinj}{\cv_{{inj}}}
\newcommand{\cvssur}{\cv_{{sur}}}
\newcommand{\cvsbij}{\cv_{{bij}}}
\newcommand{\fcvs}{\fuzz\cvs}

\newcommand{\fcvssur}{\fuzz\cvssur}
\newcommand{\fcvsbij}{\fuzz\cvsbij}
\newcommand{\isod}{IsoCluster_{\delta}}

\newcommand{\zerooneop}{(0,1]^{\text{op}}}

\newcommand{\flatten}{Flatten}

\newcommand{\loss}{\mathbf{Loss}}
\newcommand{\lm}{\mathbf{L}}
\newcommand{\floss}{\mathbf{F}\loss}

\newcommand{\mxm}{\mathbf{Mat}_{|X|,m}}

\newcommand{\mnm}{\mathbf{Mat}_{n,m}}

\newcommand{\fuzzysimplex}{\mathit{FuzzySimplex}}

\newcommand{\lk}{\mathcal{L}_k}
\newcommand{\vlk}{\mathcal{VL}_k}

\newcommand{\slink}{\mathcal{SL}}
\newcommand{\mlink}{\mathcal{ML}}

\newcommand{\mds}{\mathit{MDS}}

\newcommand{\fce}{\mathit{FCE}}

\usepackage{times}

\usepackage{tikz}
\usepackage{freetikz}
\usepackage{caption}
\usepackage{bbm}

\usepackage{algorithm,graphicx}
\usepackage[noend]{algpseudocode}

\usepackage{enumitem}
\usepackage{amsthm}
\usepackage{amsmath}
\usepackage{amssymb}
\usepackage{tikzit}

\usetikzlibrary{
  cd,
  math,
  decorations.markings,
  decorations.pathreplacing,
  positioning,
  arrows.meta,
  circuits.logic.US,
  calc,
  fit,
  trees,
  quotes}
\usetikzlibrary{decorations.pathmorphing}
\usetikzlibrary{decorations.markings}
\usetikzlibrary{decorations.pathreplacing}
\usetikzlibrary{arrows}
\usetikzlibrary{shapes.geometric}

\tikzstyle{new edge style 0}=[<-]

\newtheorem{proposition}{Proposition}

\newtheorem{definition}{Definition}[section]

\author{Dan Shiebler}
\title{Functorial Manifold Learning}
\def\titlerunning{Functorial Manifold Learning}
\def\authorrunning{{Dan Shiebler}}

\usepackage{txfonts}

\begin{document}
\maketitle

\begin{abstract}
  We adapt previous research on category theory and topological unsupervised learning to develop a functorial perspective on manifold learning, also known as nonlinear dimensionality reduction.  We first characterize manifold learning algorithms as functors that map pseudometric spaces to optimization objectives and that factor through hierarchical clustering functors.  We then use this characterization to prove refinement bounds on manifold learning loss functions and construct a hierarchy of manifold learning algorithms based on their equivariants. We express several popular manifold learning algorithms as functors at different levels of this hierarchy, including Metric Multidimensional Scaling, IsoMap, and UMAP. Next, we use interleaving distance to study the stability of a broad class of manifold learning algorithms. We present bounds on how closely the embeddings these algorithms produce from noisy data approximate the embeddings they would learn from noiseless data. Finally, we use our framework to derive a set of novel manifold learning algorithms, which we experimentally demonstrate are competitive with the state of the art. 
%   All code from the experiments can be found on GitHub at \href{https://github.com/dshieble/FunctorialManifoldLearning}{https://github.com/dshieble/FunctorialManifoldLearning}.
\end{abstract}

\section{Introduction}\label{introduction}

Suppose we have a finite pseudometric space $(X, d_X)$ that we assume has been sampled from some larger space $\bx$ according to some probability measure $\mu_{\bx}$ over $\bx$. \textbf{Manifold Learning} algorithms like Isomap \cite{tenenbaum2000global}, Metric Multidimensional Scaling \cite{abdi2007metric}, and UMAP \cite{mcinnes2018umap} construct $\rl^m$-embeddings for the points in $X$, which we interpret as coordinates for the support of $\mu_{\bx}$. These techniques are based on the assumption that this support can be well-approximated with a manifold.
% For example, a common assumption is that $d_X$ is a noisy approximation of geodesic distance on the support of $\mu_{\bx}$, and that smaller distances between points are be more likely to be reliable than larger distances. Manifold learning algorithms like Laplacian Eigenmaps, which ignore all pairs of points with distances greater than some threshold $\delta$, exploit this assumption \cite{belkin2003laplacian}. 
In this paper we use \textbf{functoriality}, a basic concept from Category Theory, to explore two aspects of manifold learning algorithms:
\begin{itemize}
    \item \textbf{Equivariance}: A manifold learning algorithm is equivariant to a transformation if applying that transformation to its inputs results in a corresponding transformation of its outputs. Understanding the equivariance of a transform lets us understand how it will behave on new types of data.
    \item \textbf{Stability}: The stability of a manifold learning algorithm captures how well the embeddings it generates on noisy data approximate the embeddings it would generate on noiseless data.  Understanding the stability of a transform helps us predict its performance on real-world applications.
\end{itemize}

% equivariants and stability of manifold learning algorithms.
% In doing so, we reveal a deep relationship between manifold learning and clustering.

\subsection{Functoriality}

In order for a manifold learning algorithm to be useful, the properties of the embeddings that the algorithm derives from $(X, d_X)$ must be somewhat in line with the structure of $(X, d_X)$. One way to formalize this is to cast the algorithm as a functor between categories. A \textbf{category} is a collection of objects and morphisms between them. Morphisms are closed under an associative composition operation, and each object is equipped with an identity morphism. An example of a category is the collection of sets (objects) and functions (morphisms) between them.

A \textbf{functor} is a mapping between categories that preserves identity morphisms and morphism composition. Underlying this straightforward definition is a powerful concept: the functoriality of a transformation is a blueprint for its structure, expressed in terms of the equivariants it preserves. If a given transformation is functorial over some pair of categories, then the transformation preserves the structure represented by those categories' morphisms.
% One way to develop a deeper understanding of an unsupervised learning algorithm is to identify the
By identifying the settings under which an algorithm is functorial, we can derive extensions of the algorithm that preserve functoriality and identify modifications that break it. See 
``Basic Category Theory'' \cite{leinster2016basic} or ``Seven Sketches in Compositionality'' \cite{fong2018seven} for more details on categories and functoriality.

\subsection{Summary of Contributions}
In Section \ref{4-manifold}, we demonstrate that manifold learning algorithms can be expressed as optimization problems defined on top of hierarchical overlapping clusterings. That is, we can express these algorithms in terms of the composition of:
\begin{itemize}
    \item a strategy for clustering points at different distance scales
    \item an order-preserving transformation from a clustering of points to a loss function
\end{itemize}
We formalize this relationship in terms of the composition of functors between categories of pseudometric spaces, hierarchical overlapping clusterings, and optimization problems. This allows us to formally extend clustering theory into manifold learning theory.

In Section \ref{spectrum} we build on clustering theory to demonstrate that every manifold learning objective lies on a spectrum based on the criterion by which embeddings are penalized for being too close together or too far apart. In Section \ref{characterization} we introduce a hierarchy of manifold learning algorithms and categorize algorithms
% as isometry-invariant, expansive-contractive, or extensible
based on the dataset transformations over which they are equivariant. In Section \ref{examples} we provide several examples of this categorization. We show that UMAP is equivariant to isometries and both IsoMap and Metric Multidimensional Scaling are equivariant to surjective non-expansive maps. Identifying these equivariants is helpful for identifying the circumstances under which we would expect our algorithms to generalize. For example, while adding points to a dataset will modify the IsoMap objective in a predictable way, this will completely disrupt the UMAP objective. This is caused by the fact that UMAP uses a local distance rescaling procedure that is density-sensitive, and is therefore not equivariant to injective or surjective non-expansive maps.

In Section \ref{stability} we use interleaving distance to study the stability of a broad class of manifold learning algorithms. We present novel bounds on how well the embeddings that algorithms in this class learn on noisy data approximate the embeddings they learn on noiseless data. 

% TODO: We also demonstrate the empirical ramifications of this characterization....
In Section \ref{recombination} we build on this theory to describe a strategy for deriving novel algorithms from existing manifold learning algorithms. As an example we derive the \textbf{Single Linkage Scaling} algorithm, which projects samples in the same connected component of the Rips complex as close together as possible. We also present experimental results demonstrating the efficacy of this algorithm.

\subsection{Related Work}
Several authors have explored functorial perspectives on clustering algorithms. Carlsson et al.\ \cite{carlsson2013classifying} introduce clustering functors that map metric spaces to partitionings, whereas Culbertson et al.\ \cite{culbertson2016consistency} take a slightly broader scope and also explore overlapping clustering functors that map metric spaces to coverings. Both approaches demonstrate that metric space categories with fewer morphisms permit a richer class of clustering functors. For example, while the single linkage clustering algorithm is functorial over the full category of metric spaces and non-expansive maps, density-sensitive clustering algorithms like robust single linkage are only functorial over the subcategory of metric spaces and injective non-expansive maps.
% This enables them to relax the requirement that the relation that groups points into clusters is transitive, which allows the clustering output to preserve richer information about the original space.
% For example, the output of the maximal linkage clustering algorithm preserves information about the nearest neighbors of each point.
In order to get around the Kleinberg Impossibility Theorem \cite{kleinberg2003impossibility}, which states that any scale invariant flat clustering must sacrifice either surjectivity or a notion of consistency, several authors \cite{carlsson2013classifying, culbertson2018functorial, scoccola2020locally} also explore hierarchical clustering functors that map metric spaces to multi-scale dataset partitionings or covers. Shiebler \cite{shieblerfunctorial} builds on this perspective to factor clustering functors through a category of simplicial complexes.

% One issue with any theory of clustering is that many important clustering algorithms are sensitive to scale and therefore may produce different clusterings for $(X, d_X)$ and $(X, 2 * d_X)$. In particular, \cite{kleinberg2003impossibility} show that any scale equivariant clustering algorithm must sacrifice either surjectivity or a notion of consistency. \cite{carlsson2013classifying} get around this restriction by defining \textbf{hierarchical clustering algorithms} that generate a series of clusterings, each at a different scale. \cite{culbertson2018functorial} also define hierarchical clustering functors that map metric spaces to multi-scale dataset partitionings.

Manifold learning shares structure with hierarchical clustering, and some authors have begun applying categorical ideas to manifold learning. For example, McInnes et al.\ \cite{mcinnes2018umap} introduce the UMAP manifold learning algorithm in terms of Spivak's fuzzy simplicial sets \cite{spivak2012metric}, which are a categorical analog of simplicial filtrations. 

In Section \ref{stability} we study the stability of manifold learning algorithms to dataset noise. Due to the importance of this topic, many other authors have researched the stability properties of manifold learning algorithms. For example, Baily \cite{bailey2012} explores adaptations of PCA to noisy data, and Gerber et al.\ \cite{gerber2007robust} demonstrate that Laplacian Eigenmaps has nicer stability properties than IsoMap. However, we believe that ours is the first work that uses interleaving distance to formalize a stability property.

% However, to the best of our knowledge there is no framework for manifold learning that is similar to \cite{carlsson2013classifying} and \cite{culbertson2016consistency}'s clustering frameworks.

\subsection{Preliminaries on Functorial Hierarchical Overlapping Clustering}\label{preliminaries}

We briefly review some definitions from the functorial perspective on hierarchical overlapping clustering. For more details, see Shiebler \cite{shieblerfunctorial}. Given a set $X$, a \textbf{non-nested flag cover} $\cl_X$ of $X$ is a cover of $X$ such that: (1) if $S_1, S_2 \in \cl_{X}$ and $S_1 \subseteq S_2$, then $S_1 = S_2$, (2) the simplicial complex with vertices corresponding to the elements of $X$ and faces to all finite subsets of the sets in $\cl_{X}$ is a flag complex, or a simplicial complex that can be expressed as the clique of its $1$-skeleton. Intuitively, $\cl_X$ is a non-nested flag cover if there exists a graph whose cliques are the sets in $\cl_X$. The category $\cvs$ has tuples $(X, \cl_{X})$ as objects where $\cl_{X}$ is a non-nested flag cover of the finite set $X$. The morphisms between $(X, \cl_{X})$ and $(Y, \cl_{Y})$ are functions $f: X \rightarrow Y$ where for any set $S$ in $\cl_{X}$ there exists some set $S'$ in $\cl_{Y}$ such that $f(S) \subseteq S'$.

Next, we will represent datasets with the category $\pmet$ of finite pseudometric spaces and non-expansive maps between them. Given a subcategory $\db$ of $\pmet$, a
\textbf{flat
$\db$-clustering functor} is a functor $C: \db \rightarrow \cvs$ that is the identity on the underlying set.
Intuitively, a flat $\db$-clustering functor maps a dataset $(X, d_X)$ to a cover of the set $X$ in such a way that increasing the distances between points in $X$ may cause clusters to separate.

A \textbf{fuzzy non-nested flag cover} is a functor $F_X: (0,1]^{op} \rightarrow \cvs$ such that for any morphism $a \leq a'$ in $(0,1]$, the morphism $F_X(a \leq a')$ is the identity on the underlying set. In the category $\fcvs$ objects are fuzzy non-nested covers and morphisms are natural transformations between them. Given a subcategory $\db$ of $\pmet$ a \textbf{hierarchical $\db$-clustering functor} is a functor $H: \db \rightarrow \fcvs$ such that for $a \in (0,1]$, $H(-)(a): \pmet  \rightarrow \cvs$ is a flat $\db$-clustering functor. Intuitively, a hierarchical $\db$-clustering functor maps a pair of a dataset $(X, d_X)$ and a strength $a \in (0,1]$ to a cover of the set $X$ in such a way that increasing the distances between points in $X$ or increasing the strength $a$ may cause clusters to separate.

\section{Manifold Learning}\label{4-manifold}%!TEX root = main.tex

% NOTE: The pairwise loss functions thing is really an explicit specification that manifold learning algorithms only care about pairwise distances. 

% NOTE: We need to use multiple loss functions - one for each pair of points, which we integrate over?

Given a choice of $m \in \mathbb{N}$, a manifold learning algorithm constructs an $n \times m$ real valued matrix of embeddings in $\mnm = \rl^{n \times m}$ from a finite pseudometric space with $n$ points. In this work we focus on algorithms that operate by solving \textbf{embedding optimization problems}, or tuples $(n, m, l)$ where $l:  \mnm \rightarrow \rl$ is a loss function, or a penalty term. We call the set of all $A \in \mnm$ that minimize $l(A)$ the \textbf{solution set} of the embedding optimization problem. In particular, we focus on \textbf{pairwise embedding optimization problems}, or embedding optimization problems where $l$ can be expressed as a sum of pairwise terms $\lij{}: \rlp \rightarrow \rl$ such that $l(A) = \sumijn \lij{}(\adistij)$. Intuitively the terms $\lij{}$ act as a penalty on bad embeddings. We express such a pairwise embedding optimization problem with the tuple $(n, m, \{ \lij{}\})$. 

Formally we define a \textbf{manifold learning problem} to be a function that maps the pseudometric space $(X, d_X)$ to a pairwise embedding optimization problem of the form $(|X|,m,\{\lij{}\})$. Note that this definition does not include any specification of how the optimization problem is solved (such as gradient descent or reduction to an eigenproblem). For example, the Metric Multidimensional Scaling manifold learning problem maps the pseudometric space $(X, d_X)$ to $(|X|,m,\{\lij{}\})$ where $\lij{}(\delta) =(d_X(x_i, x_j) - \delta)^2$. Optimizing this objective involves finding a matrix $A$ that minimizes $\sumijx (d_X(x_i, x_j) - \adistij)^2 $. That is, the Euclidean distance matrix of the rows of the optimal $A$ is as close as possible to the $d_X$ distance matrix.
% In general we can solve this problem with gradient descent, but under certain conditions on $(X, d_X)$ this problem reduces to an eigenproblem. 

If a manifold learning problem maps isometric pseudometric spaces to embedding optimization problems with the same solution set, we call it \textbf{isometry-invariant}. Intuitively, isometry-invariant manifold learning algorithms do not change their output based the ordering of $X$. A particularly useful property of isometry-invariant manifold learning problems is that they factor through hierarchical clustering algorithms.
% 
%
% \vspace{-1mm} 
\begin{proposition}\label{clusterfactor}
Given any isometry-invariant manifold learning problem $M$, there exists a manifold learning problem
% that we can express as the composition
$L \circ H$, where $H$ is a hierarchical overlapping clustering algorithm (as defined by Shiebler \cite{shieblerfunctorial}) and $L$ is a function that maps the output of $H$ to an embedding optimization problem, such that the solution spaces of the images of $M$ and $L \circ H$ on any pseudometric space $(X, d_X)$ are identical.
\end{proposition}\vspace{-1mm}
Intuitively, Proposition \ref{clusterfactor} holds because manifold learning problems generate loss functions by grouping points in the finite pseudometric space together.
% and penalizing learned embeddings from being too close together/far apart given how closely they are grouped together.
In order to use this property to adapt clustering theorems into manifold learning theorems we will introduce a target category of pairwise embedding optimization problems and replace functions with functors from $\pmet$ into this category. To start, consider the category $\lm$:
\vspace{-1mm}\begin{definition}
% NOTE: the contractive loss c goes in the same direction as the non-expansive (a non-expansive mapping on the source space increases the demands on the target space and increases the  loss)
%
%
The objects in $\lm$ are tuples $(n,\{\lij{}\})$ where $n$ is a natural number and $\lij{}: \rlp \rightarrow \rl$ is a real-valued function that satisfies $l_{i'j'}(x) = 0$ for $i' > n$ or $j' > n$.  $\lm$ is a preorder where $(n,\{\lij{}\}) \leq (n',\{\lij{}'\})$ iff for any $x \in \rlp, i,j \in\mathbb{N}$ we have $\lij{}(x) \leq \lij{}'(x)$.
% TODO: Do we need to add a functorial restriction on the fuzzy constraint too?
\end{definition}\vspace{-1mm}
Given a choice of $m$ we can view the object $(n,\{\lij{}\})$ in $\lm$ as a pairwise embedding optimization problem. However, $\lm$ is not quite expressive enough to serve as our target category. Recall the Metric Multidimensional Scaling manifold learning problem, which maps the pseudometric space $(X, d_X)$ to the pairwise embedding optimization problem $(|X|,m,\{\lij{}\})$ where $\lij{}(\delta) = (d_X(x_i, x_j) - \delta)^2$. Since $\lij{}$ does not vary monotonically with $d_X$, it is clear that this manifold learning problem is not a functor from $\pmet$ to $\lm$. However, note that we can express $\lij{}(A)$ as the sum of a term that increases monotonically in $d_X(x_i, x_j)$ and a term that decreases monotonically in $d_X(x_i, x_j)$:
\begin{gather*}
    \lij{}(\delta) =
    (d_X(x_i, x_j) - \delta)^2 = 
    \left(\delta^2 + d_X(x_i, x_j)^{2}\right) - 
    \left(2\delta d_X(x_i, x_j)\right)
\end{gather*}
We will see in Section \ref{examples} that the embedding optimization problems associated with many common manifold learning algorithms decompose similarly. We can build on this insight to create a new category $\loss$ with the following pullback:
\begin{center}
\begin{tikzcd}[column sep=0.5in,row sep=0.5in]
\loss  \arrow[dotted]{d} \arrow[dotted]{r} & \lm \arrow{d}{U}  \\
\lm^{op} \arrow{r}{U}  & \mathbb{N}
\end{tikzcd}

% \begin{tikzpicture}[>=latex]
% \node (w) at (0,0) {\(\loss\)};
% \node (x) at (0,-2) {\(\lm^{\text{op}}\)};
% \node (y) at (2,0) {\(\lm\)};
% \node (z) at (2,-2) {\(\mathbb{N}\)};
% \draw[->] (w) -- (y);
% \draw[->] (w) -- (x);
% \path[line] (x) -- node [midway,above] {U} (z);
% \path[line] (y) -- node [midway,above] {U}(z);
% \begin{scope}[shift=($(w)!.5!(z)$)]
% \draw +(-.5,0) -- +(0,0)  -- +(0,.5);
% \fill +(-.25,.25) circle (.05);
% \end{scope}
% \end{tikzpicture}

\end{center}
% \begin{gather*}
% \lm \xrightarrow{U} \mathbb{N} \xleftarrow{U} \lm^{op}
% \end{gather*}
%
where $U$ is the forgetful functor that maps $(n,\{\lij{}\})$ to $n$. Intuitively $\loss$ is a subcategory of $\lm^{op} \times \lm$ and we can write the objects in $\loss$ as tuples $(n,\{\cij{}, \eij{}\})$ where $(n,\{\cij{}, \eij{}\}) \leq (n',\{\cij{}', \eij{}'\})$ iff for any $x \in \rlp, i,j \in\mathbb{N}$ we have $\cij{}'(x) \leq \cij{}(x)$ and $\eij{}(x) \leq \eij{}'(x)$. Given a choice of $m$, each object $(n,\{\cij{}, \eij{})$ in $\loss$ corresponds to the pairwise embedding optimization problem $(n, m, \{\lij{}\})$ where $\lij{}(\delta) = \cij{}(\delta) + \eij{}(\delta)$.

Similarly to the representation of hierarchical clustering algorithms as maps into a category $\fcvs$ of functors $\zerooneop \rightarrow \cvs$, we will represent manifold learning algorithms as mapping into a category $\floss$ of functors $\zerooneop \rightarrow \loss$. The objects in
% In order to facilitate the stability analyses in Section \ref{stability}, we define manifold learning functors to map into a category of fuzzy embedding optimization problems rather than into $\loss$ directly.  
$\floss$ are functors $F: \zerooneop \rightarrow \loss$ that commute with the forgetful functor that maps $(n,\{\cij{}, \eij{}\})$ to $n$, and the morphisms are natural transformations. We call $n$ the \textbf{cardinality} of $F$.
% We will see in Section \ref{examples} that this does not significantly reduce the expressiveness of our construction:
% Recall that the preorder category $\zerooneop$ has elements in the set $(0,1]$ as objects with the order relation $\geq$ as morphisms.
%
% \vspace{-1mm}\begin{definition}
% The objects in the category $\floss$ are functors $F: \zerooneop \rightarrow \loss$ that commute with the forgetful functor that maps $(n,\{\cij{}, \eij{}\})$ to $n$. The morphisms in $\floss$ are natural transformations. We call $n$ the \textbf{cardinality} of $F$.
% \end{definition}\vspace{-1mm}
%
We can define a functor $\flatten: \floss \rightarrow \loss$ which maps the functor $F$ where $F(a) = (n, \{\cfaij, \efaij\})$ to the tuple $(n, \{\cij{}, \eij{}\})$ where:
\begin{gather*}
% NOTE: This is a functor because if each of the terms in a sum satisfies an inequality relation, the sum of the terms does as well
%
\cij{}(x) = \int_{a \in (0,1]} \cfaij(x) \ da
\qquad
\eij{}(x) = \int_{a \in (0,1]} \efaij(x) \ da
\end{gather*}
Therefore, each functor $F \in \floss$ with cardinality $n$ corresponds to the pairwise embedding optimization problem $(n,m,\{\lij{F}\})$ where $\lij{F}(\delta) = 
\int_{a \in (0,1]}  \cfaij(\delta) + \efaij(\delta) \ da$. We will call the sum of these terms, $\mathbf{l}_{F}(A)$, the $F$-loss:
\begin{gather*}
\mathbf{l}_{F}(A) =
\sumijn \lij{F}(\adistij) =
\sumijn \int_{a \in (0,1]}
\cfaij(\adistij) + \efaij(\adistij) \ da
\end{gather*}
%
% Intuitively, $\sumijn \cfaij(\adistij) + \efaij(\adistij)$ is a loss term that exists with the strength $a$, and the $F$-loss $\mathbf{l}_{F}(A) = \sumijn \lij{F}(\adistij)$ is the average loss across all strengths.
We can now give our definition of a manifold learning functor:
\vspace{-1mm}\begin{definition}
Suppose $\pmet$ is the category of pseudometric spaces and non-expansive maps and $\fcvs$ is the category of fuzzy non-nested flag covers and natural transformations (see Section \ref{preliminaries}). Then given the subcategories $\db \subseteq \pmet, \db' \subseteq \fcvs$, the composition $L \circ H: \db \rightarrow \floss$ forms a $\db$-\textbf{manifold learning functor} if $H: \db \rightarrow \db'$ is a hierarchical $\db$-clustering functor and $L: \db' \rightarrow \loss$ is a functor that maps a fuzzy non-nested flag cover with vertex set $X$ to some $F_X \in \floss$ with cardinality $|X|$.
\end{definition}\vspace{-1mm}
% NOTE: The reason why we don't need to place restrictions on H in order for (X, d_X) to map to (|X|, \{\cij, \eij\}, r) is that H already commutes with the forgetful functor
%
Intuitively a manifold learning functor $\db \xrightarrow{H} \db' \xrightarrow{L} \floss$ factors through a hierarchical clustering functor and sends $(X, d_X)$ to $F$ where $F(a) = (|X|, \{\cfaij, \efaij\})$. We will say that $M = L \circ H$ is in \textbf{standard form} if $M$ maps the one-point metric space $(\{*\}, 0)$ to some $F$ where $\cfaij(x)=\efaij(x)=0$ and $\forall \epsilon, \delta \in \rlp, H(X, d_X + \epsilon)(-\log(\delta)) \simeq H(X, d_X)(-\log(\delta + \epsilon))$. Each manifold learning functor corresponds to a manifold learning problem that maps $(X, d_X)$ to $(|X|,m,\mathbf{l}_{M(X, d_X)})$.

\subsection{A Spectrum of Manifold Learning Functors}\label{spectrum}

Recall the single and maximal linkage hierarchical overlapping clustering algorithms $\slink$ and $\mlink$. These algorithms map the pseudometric space $(X, d_X)$ to the fuzzy non-nested flag cover $(X, \cl_{X_a})$ where $\cl_{X_a}$ is respectively the connected components of the $-\log(a)$-Vietoris-Rips complex of $(X, d_X)$ and the maximally linked sets of the relation $R_a$ in which $x_1 R_a x_2$ if $d_X(x_1, x_2) \leq -\log(a)$ \cite{shieblerfunctorial, culbertson2016consistency}. 
% If a manifold learning functor factors through $\mlink$, then changing the distance between any two points in the finite pseudometric space will change exactly one term in the loss function: the term corresponding to those two points.  In contrast, if it factors through the single linkage functor $\slink$  \cite{shieblerfunctorial,culbertson2016consistency} then changing the distance between any two points in the finite pseudometric space may change either no terms in the loss function or all of the terms that correspond to pairs of points in the same connected component of the Vietoris-Rips complex. In fact,
If we apply functoriality to Proposition 6 in Shiebler \cite{shieblerfunctorial} we see:
\begin{proposition}\label{SingleMaximalRefine}
% TODO: Fix this proof
%
% NOTE: We need $M$ to be functorial over $\fcvsbij$ in order to have this hold because the morphisms from \ref{clusteringbounds} are in $\fcvsbij$. If $M$ is only functorial over a more restrictive category, then this wont hold. 
%
Suppose $\db$ is a subcategory of $\pmet$ such that $\pmetbij \subseteq \db$, $L \circ H$ is a $\db$-manifold learning functor such that $H$ is non-trivial \cite{shieblerfunctorial, culbertson2016consistency} and for all $a \in (0, 1]$, the functor $H(-)(a): \db \rightarrow \cvs$ has clustering parameter $\delta_{H,a}$. Then for $a \in (0,1]$ and $(X, d_X) \in \db$ we have maps:
\begin{gather}
    % Note that H(-)(a): Met -> Cvs, so we can compose M with it directly
    %
    (L \circ \mlink)(X,d_X)(e^{-\delta_{H,a}})
    \leq
    (L \circ H)(X,d_X)(a)
    \leq
    (L \circ \slink)(X,d_X)(e^{-\delta_{H,a}})
\end{gather}
that are natural in $a$ and $(X, d_X)$.
\end{proposition}
Intuitively, this proposition states that every manifold learning functor maps $(X, d_X)$ to a loss that is both no more interconnected than the loss that does not distinguish points within the same connected component of the Vietoris-Rips complex and no less interconnected than the loss that treats each pair of points independently.

There are many manifold learning functors that lie between these extremes. In particular, for any functor $L: \pmetinj \rightarrow \loss$ and sequence of clustering functors $\mlink ,H_1,H_2,...,H_n, \slink$ whose outputs refine each other, we can apply functoriality to form a sequence of manifold learning functors $L \circ \mlink \leq L \circ H_1 \leq ...  \leq L \circ H_n \leq L \circ \slink$. For example, consider the family $\lk$ of hierarchical overlapping clustering functors from Culbertson et al.\ \cite{culbertson2016consistency}: for $k \in \mathbb{N}$, the cover $\lk (X, d_X)(a)$ is the maximal linked sets of the relation $R_{a}$ where $xR_{a}x'$ if there is a sequence $x=x_1,x_2...,x_{k-1},x_k=x'$ in $X$ where $d(x_i,x_{i+1})\leq -\log(a)$. The functor $L \circ \lk$ therefore maps $(X,d_X)$ to a loss that distinguishes only between points whose shortest path in the Vietoris-Rips complex is longer than $k$. For $k>1$ this loss is more interconnected than $L \circ \mlink$ and less interconnected than $L \circ \slink$. This also suggests a recipe for generating new manifold learning algorithms (see Section \ref{recombination}): first express an existing manifold learning problem in the form $L \circ H$, and then form $L \circ \slink, L \circ \mlink$, or any of the functors along the spectrum $L \circ \lk$.

\subsection{Characterizing Manifold Learning Problems}\label{characterization}
% TODO: Is there anything we can say about the convexity of the manifold learning loss with respect to a category of transformations that it is invariant over 
% TODO: What happens to manifold learning algorithms that factor through excisive/non-excisive clustering functors?
% TODO: What about overlappping vs non-overlappping

% NOTE: The reason why nothing is functorial over all arrows is that nothing can be functorial over both non-injective and non-surjective maps

Similarly to how Carlsson et al.\ \cite{carlsson2013classifying} characterize clustering algorithms in terms of their functoriality over different subcategories of pseudometric spaces, we can characterize manifold learning algorithms based on the subcategory $\db \subseteq \pmet$ over which they are functorial.

% This depends on the statement "a manifold learning problem is a manifold learning functor"
We have already introduced isometry-invariant manifold learning problems. Any $\pmetisom$-manifold learning functor is isometry-invariant, and an isometry-invariant manifold learning problem is \textbf{expansive-contractive} if the loss that it aims to minimize decomposes into the sum of an expansive term $e$ that decreases as distances increase and a contractive term $c$ that increases as distances increase. Intuitively, expansive-contractive manifold learning problems use the term $e$ to push together points that are close in the original space and use the term $c$ to push apart points that are far in the original space. Any $\pmetbij$-manifold learning functor is expansive-contractive.

An expansive-contractive manifold learning problem is \textbf{positive extensible} if $c$ increases and $e$ decreases when we add new points to increase $|X|$.  If instead $c$ decreases and $e$ increases when we add new points to increase $|X|$, we say it is \textbf{negative extensible}. Intuitively, many positive-extensible manifold learning problems are minmax problems that aim to simultaneously minimize $|c|$ and maximize $|e|$. Any $\pmetsur$-manifold learning functor is positive extensible and any $\pmetinj$-manifold learning functor is negative extensible.
\begin{proposition}\label{posnegextensible}
Suppose $M$ is a standard form $\pmetsur$-manifold learning functor and $M'$ is a standard form $\pmetinj$-manifold learning functor. Then for any $(X,d_X)$ and $a \in (0,1]$ we have that $\eij{M(X, d_X)(a)}$, $\cij{M'(X, d_X)(a)}$ are non-positive and $\cij{M(X, d_X)(a)}$, $\eij{M'(X, d_X)(a)}$ are non-negative. 
% non-negative. Similarly, if $M'$ is a zero-centered $\pmetinj$ manifold learning functor, then for any $(X,d_X)$ and $a \in (0,1]$ we have that $\cij{M'(X, d_X)(a)}$ is non-positive and 
\end{proposition}
In the next section we show examples of manifold learning algorithms in each of these categories.

% \textbf{NOTE: The integral needs extra work if we are trying to apply the non-linear cutoff behavior of Laplacian eigenmaps}

\subsection{Examples}\label{examples}

\subsubsection{Metric Multidimensional Scaling ($\pmetsur$-Manifold Learning Functor)}\label{mmdssection}
% https://www.cs.toronto.edu/~hinton/csc2535/notes/lec11new.pdf
% PCA or MDS
The most straightforward strategy for learning embeddings is to minimize the difference between the pairwise distance matrix of the original space and the pairwise Euclidean distance matrix of the learned embeddings. The \textbf{Metric Multidimensional Scaling} algorithm \cite{abdi2007metric}  does exactly this. Given a finite pseudometric space $(X, d_X)$, the Metric Multidimensional Scaling embedding optimization problem is $(|X|, m, l)$ where $l(A) = \sumijn  (d_X(x_i,x_j) - \|A_i - A_j\|)^2$. When the distance matrix of the pseudometric space is double-centered (i.e. mean values of rows/columns are zero) this is the same objective that Principal Components Analysis (PCA) optimizes \cite{abdi2010principal}.
Now define $\mds: \fcvssur \rightarrow \floss$ to map the fuzzy non-nested flag cover $H: \zerooneop \rightarrow \cvsinj$ with vertex set $X$ to $F: \zerooneop \rightarrow \loss$ where $F(a) = (|X|, \{\cfaij, \efaij\}, 0)$ and:
\begin{gather*}
    % c is increasing in a because it is x^2+0 to start, and then positive and increasing after the cutoff. The constant makes sure 1/W_{ij}=C_{F_c}
    % https://www.wolframalpha.com/input/?i=plot+-+log%28x%29%2Fx+-+5+from+0+to+1
    % Since x is always positive, c is positive for small a and 0 after
    % 
    \cfaij(x)  =  \begin{cases}
        x^2 & \exists S \in H(a),\ x_i, x_j \in S\}) \\
        x^2 + 2x^2
        \left(1/W_{ij} - 1/a\right) & \text{else} 
        % The derivative of -\log(x) is -1/x, -\log(1) + \log(a) = \log(a)
    \end{cases}
    \\
    %
    % e is decreasing in a because it is 0 to start, and then negative and decreasing after the cutoff. The constant makes sure 2\log(W_ij)/W_ij=C_{F_c}
    % https://www.wolframalpha.com/input/?i=plot+-+log%28x%29%2Fx+-+5+from+0+to+1
    \efaij(x) = \begin{cases}
        0 & \exists S \in H(a),\ x_i, x_j \in S\}) \\
        % NOTE: We set 0 before a, and distance after. This causes the piecewise integral to evaluate to int_{a in (W_ij, 1]}. We need this because \log(1) = 0
        %
        \frac{2\log(W_{ij})}{W_{ij}} - \frac{2\log(a)}{a} & \text{else} 
        % The derivative of -log(x)^2 is -2*log(x)/x and - log(1)^2 + log(a)^2 = log(a)^2
    \end{cases}
\end{gather*}
where:
\begin{gather*}
    W_{ij}
    =
    \sup_{\geq 0}\{a \ |\ 
        a \in (0, 1],\ 
        \exists S \in H(a),\ 
        x_i, x_j \in S\}
\end{gather*}
\begin{proposition}\label{mmdsfunctor}
$\mds$ is a functor, and $\mds \circ \mlink$ is a $\pmetsur$-manifold learning functor that maps the finite pseudometric space $(X, d_X)$ to the Metric Multidimensional Scaling embedding optimization problem.
\end{proposition}\vspace{-1mm}

\subsubsection{IsoMap ($\pmetsur$-Manifold Learning Functor)}
For many real world datasets it is the case that the distances between nearby points are more reliable and less noisy than the distances between far away
points. The \textbf{IsoMap} algorithm \cite{tenenbaum2000global} redefines the distances between far apart points in terms of the distances between near points. Given a finite pseudometric space $(X, d_X)$, the IsoMap embedding optimization problem is $(|X|, m, l)$ where $l(A) = \sumijn  (d'_X(x_i,x_j) - \|A_i - A_j\|)^2$ such that $d'_X(x_i,x_j)$ is the length of the shortest path between $x_i$ and $x_j$ in the graph in which there exists an edge of length $d_X(x, x')$ between each pair of points $(x, x') \in X$ with $d_X(x, x') \leq \delta$.
%where either $x_2$ is a $k$-nearest neighbor of $x_1$ or $x_1$ is a $k$-nearest neighbor of $x_2$. 

\begin{proposition}\label{isomapfunctor}
% TODO: Is this actually injective or surjective? 
% 
For any $\delta \in \rlp$, there exists a hierarchical $\pmet$-clustering functor $\isod$ such that the $\pmetsur$-manifold learning functor $\mds \circ \isod$ maps the finite pseudometric space $(X, d_X)$ to the IsoMap embedding optimization problem.
\end{proposition}\vspace{-1mm}

\subsubsection{$k$-Path Scaling ($\pmetsur$-Manifold Learning Functor)}

Given a finite pseudometric space $(X, d_X)$ and $k \in \mathbb{N}$, suppose $d'_X(x_i,x_j)$ is the smallest $\delta$ such that there exists a path of $\leq k$ edges between $x_i$ and $x_j$ in the $\delta$-Vietoris Rips complex of $(X, d_X)$. Then the $\pmetsur$-manifold learning functor $\mds \circ \lk$ maps $(X, d_X)$ to the $k$-\textbf{Path Scaling} embedding optimization problem $(|X|, m, l)$, where $l(A) = \sumijx  (d'_X(x_i,x_j) - \|A_i - A_j\|)^2$.

\subsubsection{$k$-Vertex-Connected Scaling ($\pmetbij$-Manifold Learning Functor)}

For $k \in \mathbb{N}$ the hierarchical overlapping clustering functor $\vlk$ maps the finite pseudometric space $(X, d_X)$ to the fuzzy non-nested flag cover $(X, \cl_{X_a})$ where $\cl_{X_a}$ is the set of maximal $\min(|X|, k)$-vertex-connected subgraphs of the $-\log(a)$-Vietoris-Rips complex of $(X, d_X)$. Note that $\mathcal{VL}_1=\slink$ and $lim_{k \rightarrow \infty} \vlk = \mlink$. Note also that for $k > 1$ the map $\vlk$ is functorial over $\pmetinj$ but not all of $\pmet$ since a non-injective map may split a $k$-vertex-linked subgraph \cite{culbertson2016consistency}.

Now given a finite pseudometric space $(X, d_X)$ and $k \in \mathbb{N}$, suppose $d'_X(x_i,x_j)$ is the smallest $\delta$ such that there exists a $\min(|X|, k)$-vertex-connected subgraph of the $\delta$-Vietoris-Rips complex of $(X, d_X)$ that contains both $x_i$ and $x_j$.  Then the $\pmetbij$-manifold learning functor $\mds \circ \vlk$ maps $(X, d_X)$ to the $k$-\textbf{Vertex Connected Scaling} embedding optimization problem $(|X|, m, l)$ where $l(A) = \sumijx  (d'_X(x_i,x_j) - \|A_i - A_j\|)^2$. Note that unlike $\mds \circ \lk$, for $k>1$ the map $\mds \circ \vlk$ is not functorial over all of $\pmetsur$ since $\vlk$ is not functorial over $\pmetsur$. 
% and $\mds$ is not functorial over $\cvsinj$.

\subsubsection{UMAP ($\pmetisom$-Manifold Learning Functor)}

% Unlike Laplacian Eigenmaps and Metric Multidimensional Scaling,
The UMAP algorithm builds a local uber-metric space around each point in $X$, converts each local uber-metric space to a fuzzy simplicial complex, and minimizes a loss function based on a fuzzy union of these fuzzy simplicial complexes. Given a finite pseudometric space $(X, d_X)$, the UMAP embedding optimization problem is $(|X|, m, l)$ where $l$ is the fuzzy cross-entropy:
\begin{gather*}
    l(A) = \sumijx
    W_{ij}\  \log\left(\frac{W_{ij}}{e^{-\adistij}}\right)
    +
    (1-W_{ij})\  \log\left(\frac{1-W_{ij}}{1-e^{-\adistij} }\right)
\end{gather*}
and $W_{ij}$ is the weight of the fuzzy union of the $1$-simplices connecting $x_i$ and $x_j$ in the Vietoris-Rips complexes formed from the $|X|$ local uber-metric spaces $(X, d_{x_i})$ where:
\begin{gather*}
    d_{x_i}(x_j, x_k) =
    \begin{cases}
        d_X(x_j,x_k) - \min_{l=1...n} d_X(x_i, x_l) & i=j,i=k \\
        \infty & \text{else}
    \end{cases}
\end{gather*}

\begin{proposition}\label{fuzzysimplexfunctor}
There exists a hierarchical $\pmetisom$-clustering functor $\fuzzysimplex$ that decomposes into the composition of four functors that:
\begin{enumerate} 
    \item build a local uber-metric space around each point in $X$;
    \item convert each local uber-metric space to a fuzzy simplicial complex;
    \item take a fuzzy union of these fuzzy simplicial complexes;
    \item convert the resulting fuzzy simplicial complex to a fuzzy non-nested flag cover.
    %
    % \item construct a loss function based on this cover;
\end{enumerate}
\end{proposition}

\vspace{-1mm}\begin{proposition}\label{fcefunctor}
There exists a functor $\fce: \fcvsbij \rightarrow \floss$ where $\fce \circ \fuzzysimplex$ is a $\pmetisom$-manifold learning functor that maps the pseudometric space $(X, d_X)$ to the UMAP embedding optimization problem.
\end{proposition}\vspace{-1mm}
Since the UMAP distance rescaling procedure does not preserve non-expansive maps, even if a map from $(X, d_X)$ to $(X', d_{X'})$ is non-expansive, this will not necessarily be the case for all of the local uber-metric spaces $(X, d_{x_i})$ that we build from $(X, d_X)$ and $(X', d_{X'})$. For this reason $\fce \circ \fuzzysimplex$ is not functorial over $\pmetbij$.

\section{Stability of Manifold Learning Algorithms}\label{stability}
% 
% NOTE: Integral of convex function is convex https://math.stackexchange.com/questions/2795648/integral-of-a-convex-function-is-convex
% NOTE: Expansive loss decreases as distances increase and a approaches 0.
% NOTE: Contractive loss decreases as distances decrease and a approaches 1.
%
% TODO: For each manifold learning functor there exists another one in some class that is $\epsilon$-interleaved??
%
% TODO: By functoriality, if clustering preserves interleaving (or is stable) then so does the manifold learning algorithm. Are the overlapping clustering algorithms stable? 
%
% TODO: How do the stability properties we get from this proposition compare to those we get from spectral sampling bounds on PCA/Laplacian Eigenmaps?

We can use this functorial perspective on manifold learning to reason about the stability of manifold learning algorithms under dataset noise. An \textbf{$\epsilon$-interleaving} between the functors $F,G: \rlp \rightarrow \cb$ is a collection of commuting natural transformations between $F(\delta) \rightarrow G(\delta + \epsilon)$ and $G(\delta) \rightarrow F(\delta + \epsilon)$ \cite{chazal2009proximity, chazal2014persistence}. The \textbf{interleaving distance} $d_I$ between such functors is the smallest $\epsilon$ such that an $\epsilon$-interleaving exists. In order to study interleavings between functors in $\fcvs$ or $\floss$ whose domain is $\zerooneop$ rather than $\rlp$, we will say that the functors $F,G$ are $\epsilon_{*}$-interleaved when there is an $\epsilon$-interleaving between the functors $F \circ r$ and $G \circ r$ where $r(x) = e^{-x}$. We will also write $d_{I_{*}}(F, G) = d_I(F \circ r, G \circ r)$.
\vspace{-1mm}\begin{proposition}\label{manifoldstable}
% Equivalence of correspondence+distortion and matching definitions of the Gromov-Hausdorff distance https://eventuallyalmosteverywhere.wordpress.com/2015/02/17/gromov-hausdorff-distance-and-correspondences/
%
% NOTE: We don't need to impose commuting conditions here because everything commutes in Loss since it is a preorder
%
% NOTE: Total boundedness is from page 3 of https://arxiv.org/pdf/1207.3885.pdf
%
% NOTE: Gromov-Hausdorff distance is the smallest epsilon so that all of X's points are within epsilon of some point in Y and all of Y's points are within epsilon of some point in X
Given a subcategory $\db$ of $\pmet$,
% a hierarchical $\db$-clustering functor $H$ such that for all $\forall \epsilon, \delta \in \rlp, H(X, d_X + \epsilon)(-\log(\delta)) = H(X, d_X)(-\log(\delta + \epsilon))$,
a standard form $\db$-manifold learning functor $M = L \circ H$ and a pair of finite pseudometric spaces $(X,d_X),(Y,d_Y)$ such that there exists a pair of morphisms $f: (X,d_X) \rightarrow (Y,d_Y + \epsilon),g: (Y,d_Y) \rightarrow (X,d_X + \epsilon)$ in $\db$, we have $d_{I_*}(M(X, d_X), M(Y, d_Y)) \leq \epsilon$.
\end{proposition}\vspace{-1mm}
Proposition \ref{manifoldstable} is similar in spirit to previous results that use the Gromov-Hausdorff distance between metric spaces to bound the bottleneck or homotopy interleaving distances between their corresponding Vietoris-Rips complexes \cite{chazal2014persistence, bubenik2014categorification, scoccola2020locally, blumberg2017universality}. 
% However, since we only require that $L$ is functorial over $\pmetbij$ but the Gromov-Hausdorff distance considers correspondences between metric spaces that are not necessarily bijective, we use a more restrictive notion of metric space dissimilarity here.
As a special case, if $M$ is an $\pmetbij$-manifold learning functor and there exists an $\epsilon$-isometry between $(X,d_X),(Y,d_Y)$ then $d_{I_*}(M(X, d_X), M(Y, d_Y)) \leq \epsilon$.
% Now when two functors in $\floss$ are $\epsilon_{*}$-interleaved, we can use the solution to one functor's embedding optimization problem as an approximate solution to the other functor's problem. 
We can use this to prove the following:
\begin{proposition}\label{stabilitybounds}
% NOTE: The integral of an increasing function is convex https://math.stackexchange.com/questions/1318407/integral-of-an-increasing-function-is-convex
%
% NOTE: Minimum of parametric convex function is again convex https://math.stackexchange.com/questions/1407374/is-the-minimum-of-a-parametric-convex-function-convex-again?rq=1
%
% NOTE: It is totally okay and normal for |X| to be here since we are using the sum version of the loss. We can get rid of this if we use the mean/average version instead
%
Suppose we have a standard form $\pmetsur$-manifold learning functor $M$, a pair of $\epsilon$-isometric finite pseudometric spaces $(X,d_X),(Y,d_Y)$ and the matrices $A_X, A_Y$ that respectively minimize $\mathbf{l}_{M(X,d_X)}$ and $\mathbf{l}_{M(Y,d_Y)}$.
Then if:
\begin{gather*}
|\cij{M(X,d_X)(a)}(x)| \leq \frac{K_{\mathbf{c}}}{2}, |\cij{M(Y,d_Y)(a)}(x)| \leq \frac{K_{\mathbf{c}}}{2}
\end{gather*}
and:
\begin{gather*}
|\eij{M(X,d_X)(a)}(x)| \leq \frac{K_{\mathbf{e}}}{2}, |\eij{M(Y,d_Y)(a)}(x)| \leq \frac{K_{\mathbf{e}}}{2}
\end{gather*}
we have:
\begin{gather}
\mathbf{l}_{M(X,d_X)}\left(A_Y\right) \leq
\mathbf{l}_{M(X,d_X)}\left(A_X\right)
+
K_{\mathbf{c}} n^2 (1-e^{-\epsilon})
+
K_{\mathbf{e}}n^2(e^{\epsilon} - 1)
\end{gather}
If $\eij{M(X,d_X)(a)}(x)$ is constant in $x$ (such as for any $M$ that factors as $M = \mds \circ H$) we have:
\begin{gather}
\mathbf{l}_{M(X,d_X)}\left(A_Y\right) \leq
\mathbf{l}_{M(X,d_X)}\left(A_X\right)
+
K_{\mathbf{c}} n^2 (1-e^{-\epsilon})
\end{gather}
%
% If $M$ is a $\pmetinj$-manifold learning functor in which $|\cij{M(X,d_X)(a)}(x)|$ is constant in $x$ and $|\eij{M(X,d_X)(a)}(x)| \leq \frac{K}{2}, |\eij{M(Y,d_Y)(a)}(x)| \leq \frac{K}{2}$ we have:
% \begin{gather}
% %
% \mathbf{l}_{M(X,d_X)}\left(A_Y\right) \leq
% e^{2\epsilon}
% \mathbf{l}_{M(X,d_X)}\left(A_X\right)
% +
% K |X|^2 (e^{2\epsilon} + 2e^{\epsilon}-3)
% \end{gather}
%
\end{proposition}\vspace{-1mm}
For a very simple example, consider Multidimensional Scaling without dimensionality reduction. In this case $M = \mds \circ \mlink$ and $(X,d_X), (Y,d_Y)$ are each finite ordered $n$-element subspaces of $\rl^m$ with Euclidean distance. If we write the vectors in $X$ and $Y$ as matrices $A_X,A_Y \in \mnm$, then these matrices respectively minimize $\mathbf{l}_{M(X,d_X)}$ and $\mathbf{l}_{M(Y,d_Y)}$, and $\mathbf{l}_{M(X,d_X)}(A_X) = \mathbf{l}_{M(Y,d_Y)}(A_Y) = 0$. Since the function that sends the $i$th element of $X$ to the $i$th element of $Y$ must be an $\inf \{ 2\epsilon\ |\ \forall_{i} \|A_{X_i} - A_{Y_i}\| \leq \epsilon \}$-isometry, Proposition \ref{stabilitybounds} bounds the average of the squared distances between the paired rows of two matrices in terms of the largest such distance.

% NOTE: chord length: https://www.analyzemath.com/Geometry_calculators/arc_length_chord_and_area_of_a_sector_geometry_calculator.html
% NOTE: The loss function is equal to 0 because we perfectly match d'_X. we do not perfectly match d_X of course.
These bounds apply to a very general class of manifold learning algorithms, including topologically unstable algorithms like IsoMap \cite{balasubramanian2002isomap}. As an example, consider using IsoMap to project $n$ evenly spaced points that lie upon the surface of a radius $r$ circle in $\rl^2$ onto $\rl^1$. In this case $(X, d_X)$ is a finite ordered $n$-element subspace of $\rl^2$ with Euclidean distance, $M = \mds \circ \isod$ and for any matrix $A_X \in \mathbf{Mat}_{n,1}$ that consists of $n$ evenly spaced points along the real line such that $A_{X_{i+1}} - A_{X_{i}} = 2 r \sin(\frac{2 \pi}{2n})$ we have $\mathbf{l}_{M(X,d_X)}(A_X) = 0$. Now suppose that we instead apply IsoMap to a noisy view of $(X, d_X)$: a finite ordered $n$-element subspace $(Y, d_Y)$ of $\rl^2$ where $d_Y$ is Euclidean distance and $\forall_{i=1...n} d_X(X_i, Y_i) = d_Y(X_i, Y_i) = \|X_i - Y_i\| \leq \epsilon$. Then for any matrix $A_Y \in \mathbf{Mat}_{n,1}$ that minimizes $\mathbf{l}_{M(Y,d_Y)}$, Proposition \ref{stabilitybounds} bounds the average squared difference between $|A_{Y_{i+1}} - A_{Y_{i}}|$ and $2 r \sin(\frac{2 \pi}{2n})$.

\section{Experiments in Functorial Recombination}\label{recombination}
One benefit of the functorial perspective on manifold learning is that it provides a natural way to produce new manifold learning algorithms by recombining the components of existing algorithms. Suppose we are able to express two existing manifold learning algorithms $M_1,M_2$ in this framework such that $M_1 = L_1 \circ H_1$ and $M_2 = L_2 \ \circ H_2$ where $H_1, H_2$ are hierarchical clustering functors. Then we can use the compositionality of functors to define the manifold learning algorithms $L_2 \circ H_1$ or $L_1 \circ H_2$. We can use this procedure to combine the strengths of multiple algorithms in a way that preserves functoriality (and therefore also stability by Proposition \ref{stabilitybounds}).  For example, if we compose the  $\fuzzysimplex$ functor from Proposition \ref{fuzzysimplexfunctor} with $\mds$ we form the  $\pmetisom$-manifold learning functor $\mds \circ \fuzzysimplex$ which maps $(X, d_X)$ to the embedding optimization problem $(|X|, m, l)$ where $l(A) = \sumijx (-\log(\alpha_{ij}) - \|A_i - A_j\|)^2$ and $\alpha_{ij}$ is the strength of the fuzzy simplex that UMAP forms between $x_i$ and $x_j$.

% one of the largest contributors to UMAP's practical effectiveness is the density-aware local metric rescaling. This counteracts the non-uniformity of the data distribution, and is captured in the $\fuzzysimplex$ functor \cite{mcinnes2018umap}. Now define . I

% uses the same distance rescaling as UMAP and the same objective as Laplacian Eigenmaps. This functor corresponds to a novel manifold learning algorithm, which we call \textbf{L-UMAP} (Algorithm \ref{lumappseudo}). This algorithm benefits from both the locality-preserving character of Laplacian Eigenmaps and the robustness of UMAP's density-based distance rescaling. We can solve the embedding optimization problem that L-UMAP forms for a dataset $(X, d_X)$ by computing the eigenvectors of the weighted Laplacian of the fuzzy simplicial complex that UMAP would construct from $(X, d_X)$. 
% % $\lehd \circ \fuzzysimplex$ uses the same distance rescaling as UMAP and the same objective as Laplacian Eigenmaps.
% % Minimizing the loss function formed by composing this functor with $\lehd$ is roughly equivalent to computing the eigenvectors of the weighted Laplacian of UMAP's fuzzy simplicial complex \citep{belkin2003laplacian}. That is, 
% We demonstrate in Table \ref{lumapresults} that a model trained on L-UMAP embeddings performs competitively with a model trained on either Laplacian Eigenmaps or UMAP embeddings on the Fashion MNIST dataset \citep{xiao2017/online} and the 20 Newsgroups \citep{lang1995newsweeder} classification datasets.

\begin{figure}
\includegraphics[width=16cm,height=8cm]{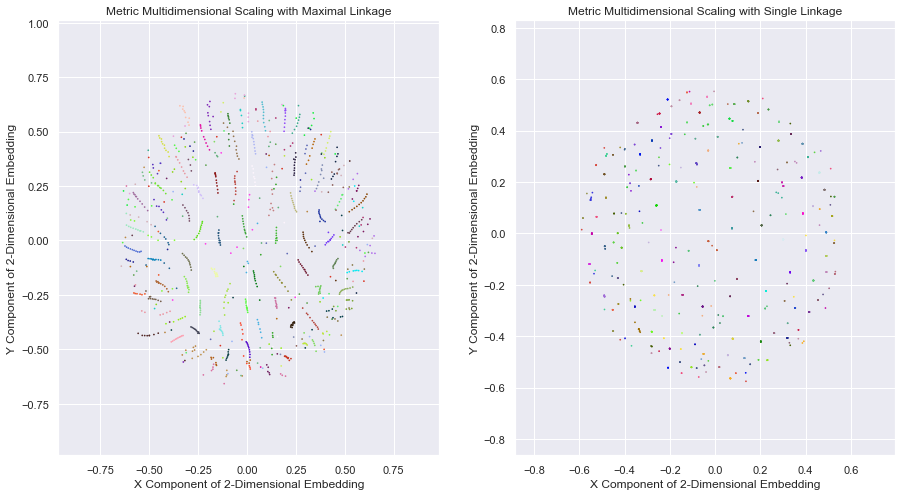}
\caption{Embeddings of DNA sequences from the DNA recombination task with $L=1000,N=100,M=10$.
% of each DNA sequence is $1000$, the number of mutation lists $N$ is $100$, and the length of each mutation list $M$ is $10$.
Each color indicates a unique DNA sequence mutation list. Note that Single Linkage Scaling ($\mds \circ \slink$) on the right embeds sequences in the same mutation list more closely together than Metric Multidimensional Scaling ($\mds \circ \mlink$) on the left.}
\label{mutationsfigure}
\end{figure}

For a more illustrative example, consider a DNA recombination task in which we attempt to match a string of DNA that has been repeatedly mutated back to the original string. One way to solve this task is to generate embeddings for each string of DNA and look at the nearest neighbors of the mutated string. We can simulate this task as follows
\begin{enumerate}
    \item Generate $N$ original random sequences of DNA of length $L$ (strings of ``A'', ``C'', ``G'', ``T'').
    \item For each sequence, mutate the sequence $M$ times to produce a mutation list, or a list of sequences which each start with an original DNA sequence and end with a final DNA sequence.
    % There will be $N$ total mutation lists, each of length $M$.
    %
    \item Collect each of the $M$ sequences in each of these $N$ mutation lists into an $N*M$ element finite pseudometric space with Hamming distance.
    \item Build embeddings from this pseudometric space and compute the percentage of mutation lists for which the nearest neighbor of the last DNA sequence in that list among the set of all original sequences is the first sequence in that list (the accuracy).
\end{enumerate}
A manifold learning algorithm that performs well on this task will need to take advantage of the intermediate mutation states to recognize that the first state and final state in a mutation list should be embedded as close together as possible. We can follow the procedure in Section \ref{spectrum} to adapt the Metric Multidimensional Scaling algorithm $\mds \circ \mlink$ (Section \ref{mmdssection}) into such an algorithm by forming the maximally interconnected functor $\mds \circ \slink$.
Intuitively, this functor maps $(X, d_X)$ to a loss function that corresponds to the optimization objective for Metric Multidimensional Scaling where Euclidean distance is replaced with:
\begin{gather*}
d^{*}_X(x,x') = \inf\{
    \delta \ |\ \exists x=x_1,x_2,...,x_n=x' \in X,
    d_X(x_i,x_{i+1}) \leq \delta
    \}
\end{gather*}
We call this the \textbf{Single Linkage Scaling} algorithm (Algorithm \ref{mdsslinkpseudo}).
Since this algorithm embeds points that are connected by a sequence in the original space as close together as possible, we expect Single Linkage Scaling to outperform Metric Multidimensional Scaling on this task. This is exactly what we see in Table \ref{dnarecombinationtable}. We also show the embeddings for each sequence in each list in Figure \ref{mutationsfigure}.
% All code from the experiments in this section can be found on GitHub at \href{https://github.com/dshieble/FunctorialManifoldLearning}{https://github.com/dshieble/FunctorialManifoldLearning}.
%

\begin{algorithm}
  \caption{Single Linkage Scaling}\label{mdsslinkpseudo}
  \begin{algorithmic}[1]
    \Procedure{SingleLinkageScaling}{($(X, d_X), m$)}
      \State Initialize the $|X| \times |X|$ matrix $B$ to all zeros
      \State $\forall i,j \leq |X|$
      \State \indent $B_{ij} = \inf\{
            \delta \ |\ \exists x_i=x_1,x_2,...,x_n=x_j \in X,
            d_X(x_k,x_{k+1}) \leq \delta
            \}$
      \State $A \gets \min_{A \in \mxm} \sumijx (\adistij - B_{ij})^2$ 
      \State \textbf{return} $A$
      %\Comment{The g.c.d.\ is $b$}
    \EndProcedure
  \end{algorithmic}
\end{algorithm}

\begin{table}
    \centering
    % \begin{adjustwidth}{-.75cm}{}
    \begin{tabular}{|c|c|c|c|c|}
        \hline
        \multicolumn{1}{|p{5.5cm}|}{\centering Algorithm} &
        \multicolumn{1}{|p{2cm}|}{\centering  Accuracy \\ $N=100$ \\ $M=10$}
        &
        \multicolumn{1}{|p{2cm}|}{\centering  Accuracy \\ $N=100$ \\ $M=20$}
        &
        \multicolumn{1}{|p{2cm}|}{\centering  Accuracy \\ $N=200$ \\ $M=10$}
        &
        \multicolumn{1}{|p{2cm}|}{\centering  Accuracy \\ $N=200$ \\ $M=20$}
        \\
        %[0.5ex]
        \hline
        \hline
        
        \multicolumn{1}{|p{5.5cm}|}{\centering Metric Multidimensional Scaling \\ Embedding Size 2} & 0.21 ($\pm$ 0.05) & 0.01 ($\pm$ 0.02) & 0.29 ($\pm$ 0.02) & 0.01 ($\pm$ 0.00) \\
        \hline
        \multicolumn{1}{|p{5.5cm}|}{\centering Single Linkage Scaling \\ Embedding Size 2} & 0.61 ($\pm$ 0.02) & 0.68 ($\pm$ 0.05) & 0.76 ($\pm$ 0.01) & 0.32 ($\pm$ 0.02) \\
        \hline
        \multicolumn{1}{|p{5.5cm}|}{\centering Metric Multidimensional Scaling \\ Embedding Size 5} & 0.74 ($\pm$ 0.01) & 0.13 ($\pm$ 0.02) & 0.84 ($\pm$ 0.01) & 0.04 ($\pm$ 0.01) \\
        \hline
        \multicolumn{1}{|p{5.5cm}|}{\centering Single Linkage Scaling \\ Embedding Size 5} & 0.93 ($\pm$ 0.05) & 0.91 ($\pm$ 0.02) & 0.96 ($\pm$ 0.02) & 0.34 ($\pm$ 0.02) \\
        \hline
    \end{tabular}
    % \end{adjustwidth}
    \caption{Accuracy on the DNA recombination task of the Metric Multidimensional Scaling ($\mds \circ \mlink$) and Single Linkage Scaling ($\mds \circ \slink$) algorithms (higher numbers are better). The accuracy is the percentage of the $N$ mutation lists of length $M$ for which the nearest neighbor of the last sequence in the list among the set of all original DNA sequences is the first sequence in that list. The reported numbers are means (and standard deviations) across $10$ simulations. All DNA sequences are of length $L=1000$.}
\label{dnarecombinationtable}
\end{table}

\section{Discussion and Future Work }\label{5-discussion}%!TEX root = main.tex

We have taken the first steps towards a categorical framework for manifold learning. By defining an algorithm as a functor from a category of metric spaces, we can explicitly express the kind of dataset transformations under which it is equivariant. We have shown that for many popular manifold learning algorithms, including Metric Multidimensional Scaling and IsoMap, the optimization objective changes in a predictable way as we modify the metric space. 

The functorial perspective also suggests a new strategy for exploratory data analysis with manifold learning. Since we can decompose manifold learning algorithms into two components (clustering and loss) we can examine how slight variations of the clustering algorithm affect the learned embeddings. We have shown in Section \ref{spectrum} that every manifold learning functor $L \circ H$ lies on a spectrum of interconnectedness between $L \circ \mlink$ and $L \circ \slink$, and we can form new algorithms by moving along this spectrum. For example, we saw in Section \ref{recombination} that replacing the $\mlink$ functor with $\slink$ in the Metric Multidimensional Scaling algorithm substantially changes the learned embeddings and improves performance on a DNA recombination task. There are also many algorithms that lie between these two options, including the $k$-Path Scaling and $k$-Vertex-Connected Scaling algorithms that we introduced in Section \ref{examples}. 

Another major benefit of expressing algorithms as functors is that functors preserve categorical properties like interleaving distance. This allows us to easily reason about the stability properties of both existing algorithms and new algorithms that we create by recombining functors. Other authors have used this strategy to prove stability properties of the homology of geometric filtered complexes \cite{chazal2014persistence}. In Section \ref{stability} we have used this strategy to define bounds on how dataset noise affects optimization quality. In future work we hope to use these techniques to derive more powerful theorems around the resilience of other kinds of unsupervised or supervised algorithms to noise. For example, we may also be able to tighten our bounds by switching our perspective from finite metric spaces to distributions \cite{brown2020probabilistic} or even involving categorical probability \cite{fritz2020synthetic} to replace interleaving distance with a probabilistic analog. Due to the simplicity and flexibility of this strategy, other researchers have begun to develop more flexible characterizations of interleaving distance that we can apply in even more situations \cite{scoccola2020locally}.

\bibliographystyle{eptcs}
\bibliography{generic}

\end{document}